\documentclass[10pt,twocolumn,letterpaper]{article}

\usepackage{cvpr}              % To produce the CAMERA-READY version
\usepackage{graphicx}
\usepackage{amsmath}
\usepackage{amssymb}
\usepackage{booktabs}
\usepackage{multirow}

\usepackage[pagebackref,breaklinks,colorlinks]{hyperref}

\usepackage[capitalize]{cleveref}
\crefname{section}{Sec.}{Secs.}
\Crefname{section}{Section}{Sections}
\Crefname{table}{Table}{Tables}
\crefname{table}{Tab.}{Tabs.}

\def\cvprPaperID{*****} % *** Enter the CVPR Paper ID here
\def\confName{CVPR}
\def\confYear{2022}

\begin{document}

\title{NeReF: Neural Refractive Field for Fluid Surface Reconstruction and Implicit Representation}
\author{
Ziyu Wang${^1}$\\
% ShanghaiTech University\\
% \thanks{\tt\small wangzy6@shanghaitech.edu.cn}
\and
Wei Yang${^2}$\\
% Huazhong University of Science and Technology\\
% {\tt\small weiyangcs@hust.edu.cn}
\and
Junming Cao${^3}$\\
% Shanghai Advanced Research Institute\\
% {\tt\small caojm@sari.ac.cn}
\and
Lan Xu${^1}$\\
% ShanghaiTech University\\
% {\tt\small xulan1@shanghaitech.edu.cn}
\and
Junqing Yu${^2}$\\
% Huazhong University of Science and Technology\\
% {\tt\small yjqing@hust.edu.cn}
\and
Jingyi Yu${^1}$\\
% ShanghaiTech University\\
% {\tt\small yujingyi@shanghaitech.edu.cn}
\and
${^1}$ShanghaiTech University
\and
${^2}$Huazhong University of Science and Technology
\and
${^3}$Shanghai Advanced Research Institute
% \thanks{Email: \\
% \hspace{2mm}\indent\tt\small wangzy6@shanghaitech.edu.cn\\
% \hspace{2mm}\indent\tt\small weiyangcs@hust.edu.cn\\
% \hspace{2mm}\indent\tt\small caojm@sari.ac.cn\\
% \hspace{2mm}\indent\tt\small xulan1@shanghaitech.edu.cn\\
% \hspace{2mm}\indent\tt\small yjqing@hust.edu.cn\\
% \hspace{2mm}\indent\tt\small yujingyi@shanghaitech.edu.cn\\
% }
}
% \author{
% \author[*]{Ziyu Wang}
% \author[+]{Wei Yang}
% \author[*]{Junming Cao}
% \author[*]{Lan Xu}
% \affil[*]{ShanghaiTech University, \authorcr Email: \{wangzy6, xulan1, yujingyi\}@shanghaitech.edu.cn}
% % \affil[+]{单位2, 作者2的邮箱}
% }

\maketitle

%%%%%%%%% ABSTRACT
\begin{abstract}
Existing neural reconstruction schemes such as Neural Radiance Field (NeRF) are largely focused on modeling opaque objects. We present a novel neural refractive field (NeReF) to recover wavefront of transparent fluids by simultaneously estimating the surface position and normal of the fluid front. Unlike prior arts that treat the reconstruction target as a single layer of the surface, NeReF is specifically formulated to recover a volumetric normal field with its corresponding density field. A query ray will be refracted by NeReF according to its accumulated refractive point and normal, and we employ the correspondences and uniqueness of refracted ray for NeReF optimization. We show NeReF, as a global optimization scheme, can more robustly tackle refraction distortions detrimental to traditional methods for correspondence matching. Furthermore, the continuous NeReF representation of wavefront enables view synthesis as well as normal integration. We validate our approach on both synthetic and real data and show it is particularly suitable for sparse multi-view acquisition. We hence build a small light field array and experiment on various surface shapes to demonstrate high fidelity NeReF reconstruction. 
\end{abstract}
%%%%%%%%% BODY TEXT
\section{Introduction}
\label{sec:intro}

Existing neural reconstruction schemes such as Neural Radiance Field (NeRF) have shown its great power for scene representation and reconstruction. However, NeRF and its successors largely focus on modeling opaque objects. In contrast, modeling and reconstruction of refractive surfaces from photographs have great importance for applications ranging from fluid physics analysis, environmental monitoring to computer graphics. Refractive surfaces poses exceptional challenges for NeRF representation as a light ray only diverts from its straight path when traversing the air-fluid interface and hence makes the sampling and radiance representation of rays no longer applicable.

\begin{figure}
    \centering
    \includegraphics[width=0.98\linewidth]{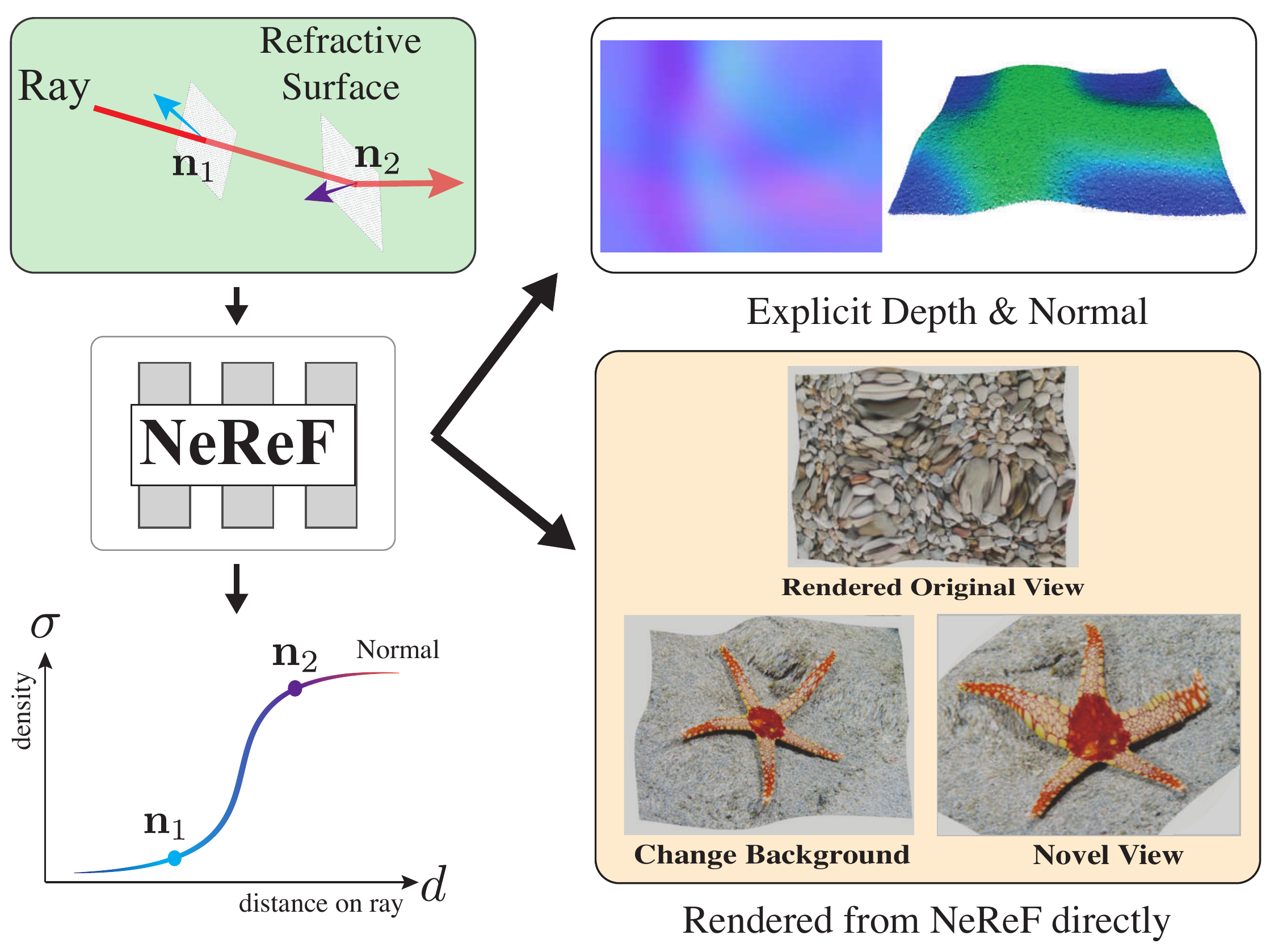}
    \caption{Our Neural Refractive Field (NeReF) is formuated to recover a volumetric normal field with its corresponding density field. Despite the explicit depth and normal representation can be derived from NeReF, we can render refraction effect and synthesize novel views directly from NeReF.}
    \label{fig:teaser}
\end{figure}

A common non-intrusive approach for estimating the shape of fluids is to analyze the distortions of a reference pattern placed under the fluid~\cite{murase1992surface}. In particular, many approaches rely on imposing additional assumptions, such as pattern appearance, water height~\cite{morris2011dynamic, qian2017stereo, shan2012refractive} and optics~\cite{han2018dense, ji2013reconstructing, ye2012angular}, while others create dedicated imaging/optics systems (e.g., camera array~\cite{ding2011dynamic}, Bokode~\cite{ye2012angular} and light field probe~\cite{wetzstein2011hand}) for acquiring fluid structures. Morris et al.~\cite{morris2011dynamic} introduce a novel refractive disparity for water surface recovery using ``stereo matching''. Qian et al.~\cite{qian2017stereo} use a camera array to estimate both water surface and the underwater scene through exploiting the surface normal consistence. Xiong and Heidrich~\cite{xiong2021wild} propose a novel differentiable framework to reconstruct the 3D shape of underwater environments from a single, stationary camera placed above the water. Notably, Thapa et al.~\cite{thapa2020dynamic} proposes learning-based single-image approach with recurrent layers modeling spatio-temporally consistence to recover dynamic fluid surfaces.

\begin{figure}
    \centering
    \includegraphics[width=0.98\linewidth]{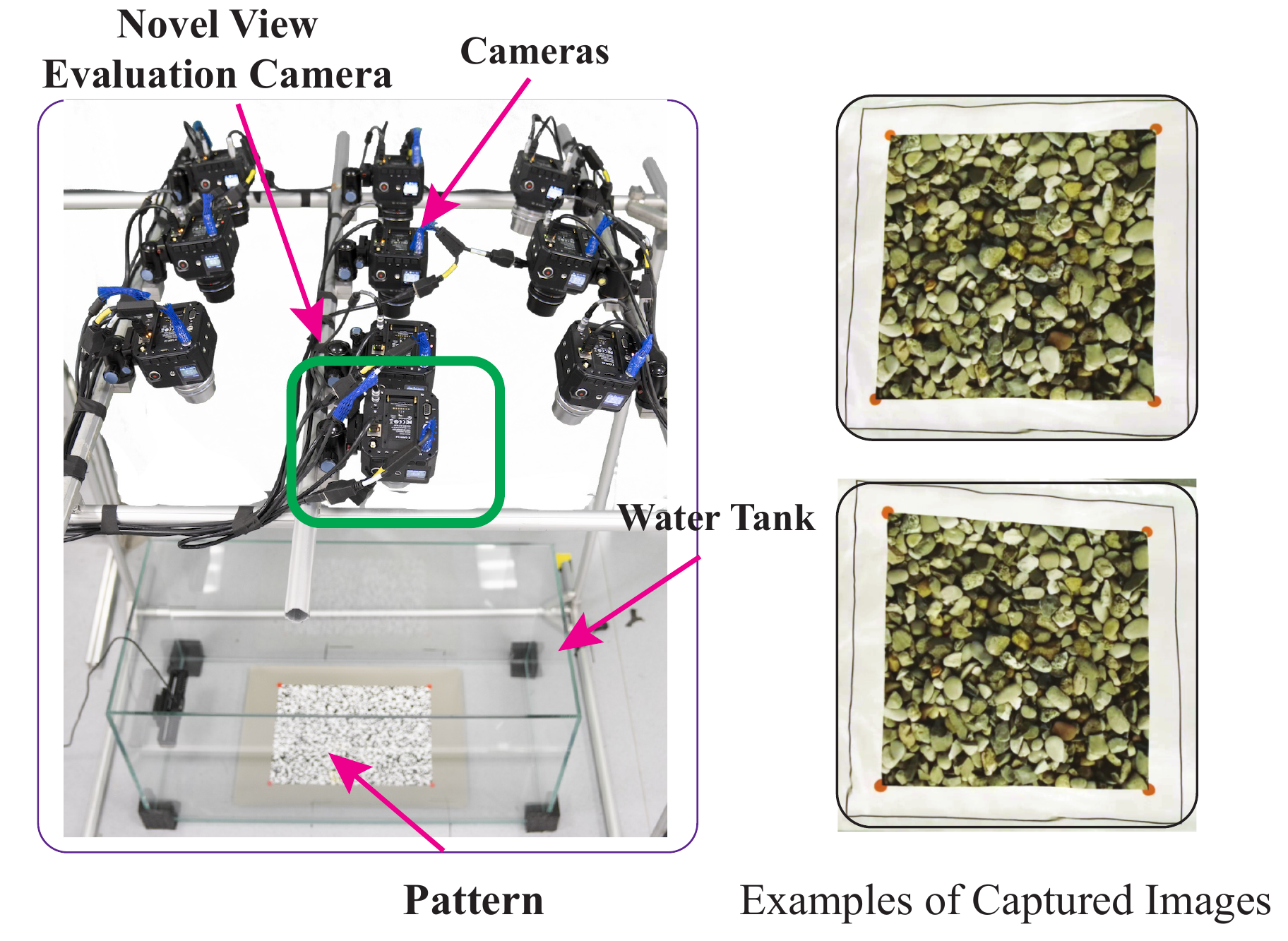}
    \caption{The real fluid surface capture system which consists of 10 Z Cam E2, a water tank and a pattern under the water. We reserve one camera specifically for the evaluation purpose. On the right shows example real images captured.}
    \label{fig:real_setup}
\end{figure}

Existing approaches unanimously model the air-fluid interface as one layer of geometry surface, which usually is explicitly represented by depth and normal maps. Re-rendering of the refraction effect from depth and normal requires ray tracing with hit point testing and two-bounce recursion. In this paper, we propose a Neural Refractive Field (NeReF) to implicitly represent a fluid surface, by taking the advantage of the recent success in neural implicit scene representation. Our work is inspired by the Neural Radiance Field (NeRF). More specifically, we use a fully-connected deep network to represent the fluid surface, whose input is a 3D coordinate and outputs the volume density and normal at the coordinate. We can perform re-rendering of refraction effects directly from the implicit representation by integrating normal along a target ray and deflecting it according to Snell's law. Explicit representations, such as depth and normal, can still be recovered effectively via volume integration (Fig.~\ref{fig:teaser}). In contrast to supervised approaches~\cite{ye2012angular}, NeReF optimization is performed in a per-scene manner and hence avoids the lack of real dataset problem. Moreover, NeReF models a continuous function and generates results at resolutions on demand. And NeReF model is relatively smaller (about 4MB), the advantage expands as the desired resolution increases. View synthesis from NeReF is easier compared to ray tracing. To evaluate our proposition, we construct a multi-camera system similar to~\cite{qian2018simultaneous} and capture a known pattern under dynamic fluid surface. We use the optical flow~\cite{murase1992surface} technique to obtain the groundtruth point-ray correspondences for NeReF optimization. Experiment results show our training free approach can recover the fluid surface with high fidelity. Our system setup is shown in Fig.~\ref{fig:real_setup}.

Compared to the original NeRF which models radiance along rays, we demonstrate that the geometric information (normal of refractive surface) of the ray can also be implicitly encoded in a neural field. %To the best of our knowledge, we are the first to demonstrate this property. 

%-------------------------------------------------------------------------
\section{Related Work}
\label{sec:related_work}

Our work is closely related to researches in fluid surface reconstruction and neural scene representation.

\begin{figure}
    \centering
    \includegraphics[width=0.98\linewidth]{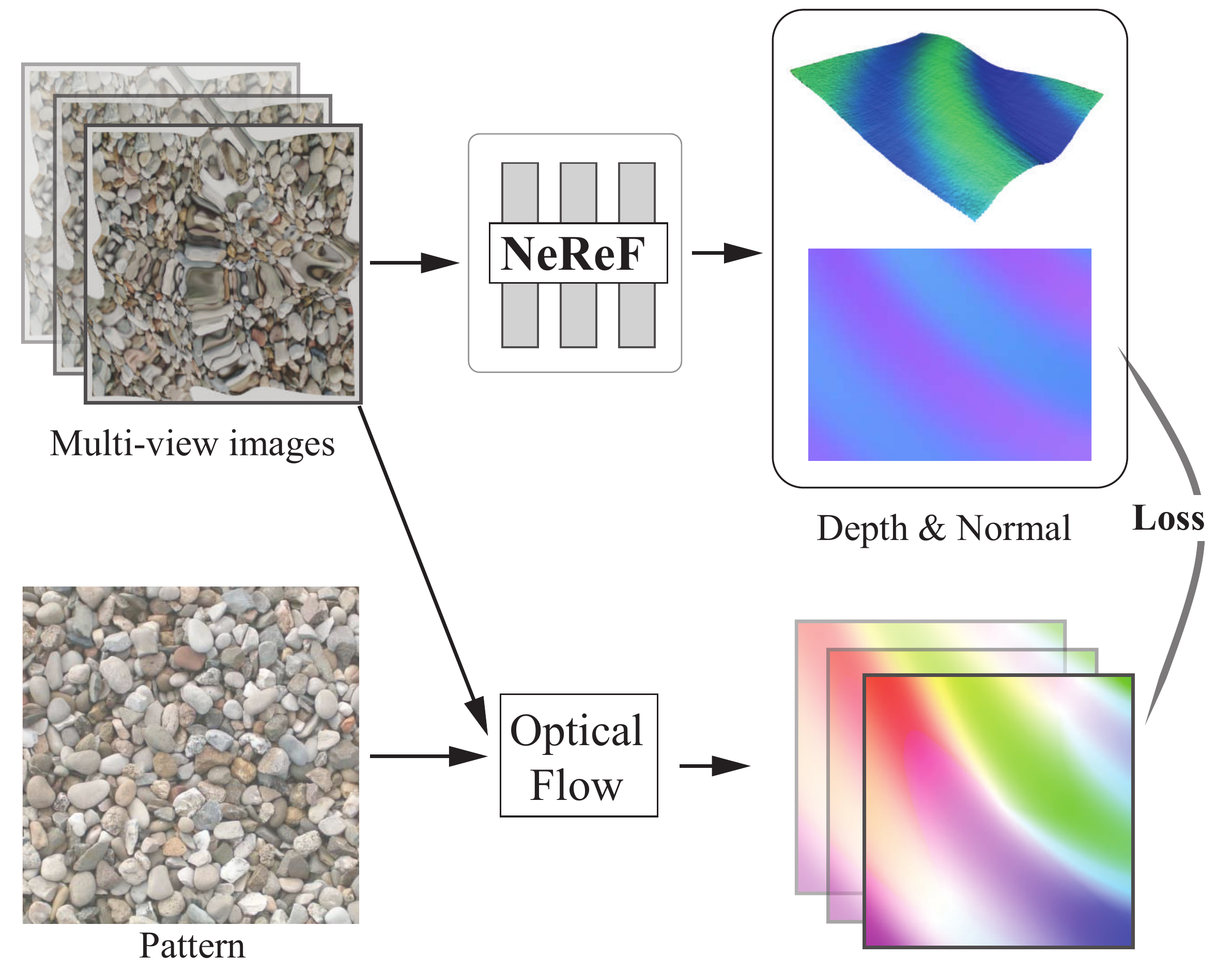}
    \caption{The overview of our fluid surface reconstruction approach based on NeReF. Specifically, given multi-view inputs of fluid surface, we first calculate the optical flow w.r.t. the patter without water. Then we train NeReF with input images and it can produce depth and normal via volumetric rendering. Then the loss is computed between recovered depth normal and optical flow with refraction physics.}
    \label{fig:overview}
\end{figure}

\textbf{Image based Fluid Surface Reconstruction} Shape from distortion analyzes the distortions in images of patterns placed under water, and it is the most common technique used for image based fluid reconstruction initiated by Murase’s pioneering work~\cite{murase1992surface}. The following shape-from-distortion methods can be categorized into single-view based methods or multi-view based methods. Approaches that adopt a single viewpoint setup usually assume additional surface constraints, such as planarity~\cite{asano2016shape, chang2011multi, ferreira2005stereo} and integrability~\cite{wetzstein2011hand, ye2012angular}, to tackle the depth normal ambiguity. Notably, Tian and Narasimhan~\cite{tian2010globally} develop a data-driven iterative algorithm to rectify the water distortion and recover water surface through spatial integration. Shan et al.~\cite{shan2012refractive} estimate the surface height map from refraction images with global
optimization. Xiong and Heidrich~\cite{xiong2021wild} propose a novel differentiable framework to reconstruct the 3D shape of underwater environments from a single, stationary camera placed above the water in the wild environment. In contrast, multi-view based approaches rely on dedicately designed imaging/optic systems. Ye~\cite{ye2012angular} exploit Bokode - a computational optical device that emulates a pinhole projector - to capture ray-ray correspondences, which can be used to directly recover the surface normals. Morris et al.\cite{morris2011dynamic} extend the traditional multi-view triangulation to be appropriate for refractive scenes, and build up a stereo setup for water surface recovery. More recently, a learning-based single-image approach has recently been presented for recovering dynamic fluid surfaces~\cite{ye2012angular}. Our capturing system setup is mainly similar to that of Qian et al.\cite{qian2017stereo}, which creates a 3x3 camera array and use optical flow technique to recover under water object(Fig.~\ref{fig:overview}). However, like other approaches, it models the air-fluid interface as depth and normal maps. In contrast, We propose to implicitly encode the fluid surface into a fully connected neural network. 

There is another line of works which aim to recover transparent objects, such as gas flow~\cite{atcheson2008time}, which we recommend to read for extra information.

\textbf{Neural Radiance Field} The remarkable work of Neural Radiance Field is a milestone in novel view synthesis area. Before Mildenhall et at~\cite{mildenhall2020nerf} raise the idea of NeRF, several methods are proposed to predict photo-realistic novel views of scenes base on dense sampling views. Theses methods can be divided into two classes. One class of methods use mesh-based representations of scenes with diffuse~\cite{waechter2014let} or view-dependent~\cite{wood2000surface, debevec1996modeling, buehler2001unstructured} appearance, optimized by differentiable rasetizers~\cite{chen2019learning, genova2018unsupervised, liu2019soft, loper2014opendr} or pathtracers~\cite{nimier2019mitsuba, li2018differentiable}. Another class of methods use volumetric representations to address this task. NeRF combines the implict representation with volumetric rendering to achieve compelling novel view synthesis with rich view-dependent effects. As the headstone of many following works, including ours, NeRF uses the weights of a multilayer perceptron(MLP) to represent a scene as a continuous volumetric field of particles that block and emit light. It takes single continuous 5D coordinates (spatial location $(x, y, z)$ and view direction $(\theta, \phi))$ as input and outputs the volume density $\sigma$ and view-dependent color $c$. 

Based on the pipeline of NeRF, several extended works have been proposed. Ricardo et al~\cite{martin2021nerf} present an approach to enable the NeRF capable of modeling uncontrolled images from unstructured photo collections with learning a per-image latent embedding apperance variations and decomposing scenes into image-dependent components. In Chen et al's~\cite{chen2021mvsnerf} work, the MVSNeRF is a deep neural network that is able of utilizing three nearby input views via fast network inference to reconstruct radiance fields. Jonathan et al~\cite{barronmip} combine the mip-map approach and NeRF together, simultaneously improve the original positional encoding into an integrated positional encoding, which represent the volume covered by each conical frustum, ending up with mip-NeRF which reduces aliasing and improves NeRF's ability to represent fine details. In Alex Yu et al~\cite{yu2021plenoctrees}'s PlenOctreees for real-time rendering of Neural radiance fields, spherical harmonics are applied as the base of a representation of view-dependent colors. Besides, PlenOctree is applied to store density and SH coefficients modelling view-dependent appearance at each leaf. With these two measures, this method can render images at more than 150FPS, which is thousands times faster than the conventional method. 

\textbf{Neural Scene Representation} Scene representation is a process that interprets the visual data into a feature representation. By providing the aimed pose and latent code~\cite{sitzmann2019deepvoxels} or alternatively transform views directly in the latent space\cite{worrall2017interpretable}, novel view images can be rendered. Generative Query Network(GQN)~\cite{eslami2018neural} provides framework within which machine learning scenes using their own sensors. However, this framework has the limitation of being oblivious to the 3D structure. Voxel grid representations and graph neural networks are the method that capture 3D structures. Vincent Sitzmann et al~\cite{sitzmann2019scene} propose a continuous 3D-structure-aware neural scene representation network, which trained from 2D images and their camera poses and encodes both geometry and appearance. Following their previous work, Sitzmann propose the light field network. This neural network represents the light field of a 3D scene implicitly, which can be used to extract depth maps from 360-degree light fields.

\section{Neural Refractive Field}
%Neural refraction field

The core at our fluid surface reconstruction approach is as a \textbf{ne}ural \textbf{re}fraction \textbf{f}ield (NeReF), which retains the continuous scene representation ability of NeRF. Before proceeding, let's briefly review NeRF for easier explanation. NeRF represents a scene using a fully-connected deep network, whose input is a spatial location and viewing direction and output is the volume density and view-dependent emitted radiance. A novel image view is synthesized by sampling coordinates along camera rays and use volume rendering techniques to project the output colors and densities into an image. In this paper, we use the same scheme but adapt the NeRF for generating a normal value at each spatial location.

\begin{figure*}
    \centering
    \includegraphics[width=1.0\linewidth]{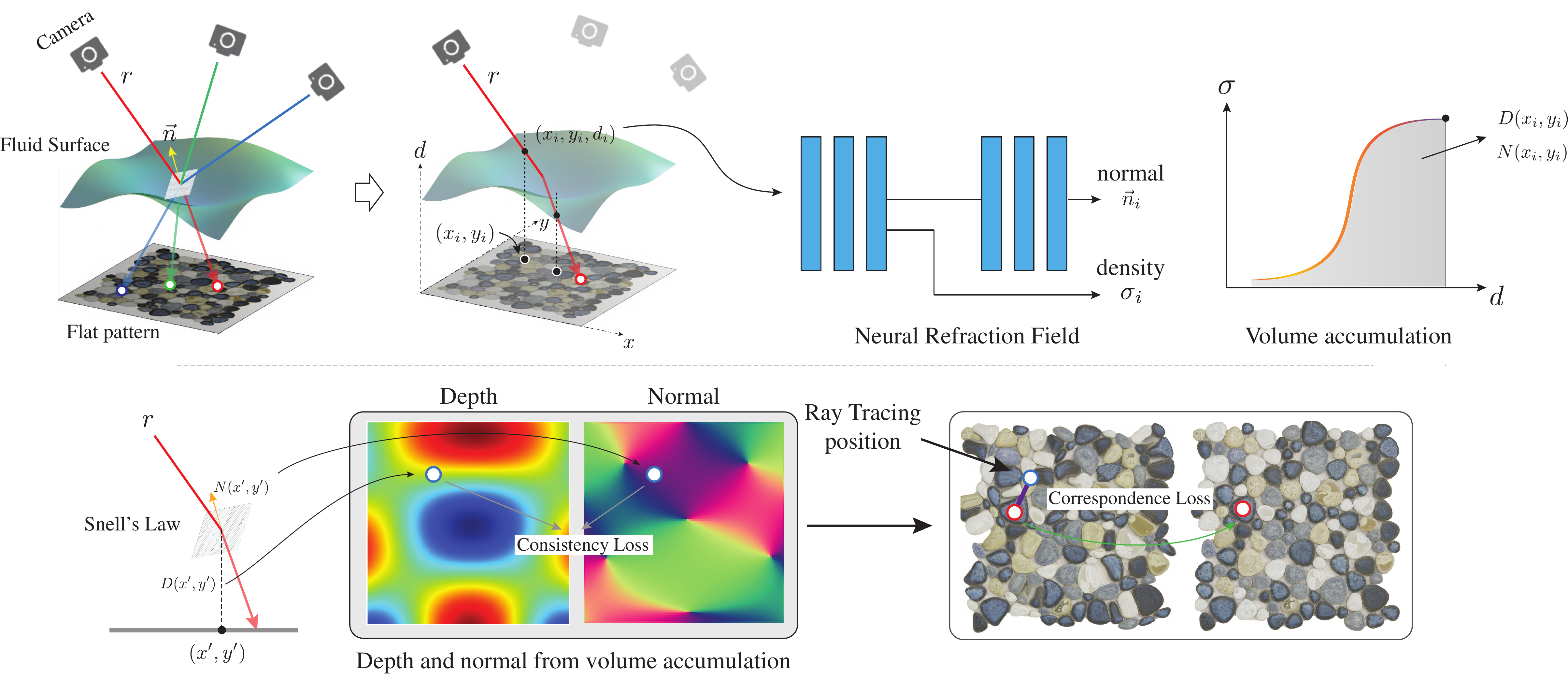}
    \caption{The pipeline of our fluid surface recontruction method from NeReF. A sample point on input ray goes through a MLP network to predict the density and normal. The depth and normal can be explicitly rendered from NeReF and we use a correspondence and depth normal consistency loss to optimize NeReF.}
    \label{fig:pipeline}
\end{figure*}

Specifically, we represent the continuous fluid surface as an implicit function $\mathcal{F}$, which is approximated by a Multi-Layer Perception (MLP).  It takes a 3D location $(x, y, z)$ as input, and outputs the surface normal $\mathbf{n}$ along with the volume density $\sigma$.
\begin{equation}
    \mathcal{F}:(x, y, z) \rightarrow (\sigma, \mathbf{n})
\end{equation}

Note that although this modification is relatively small, the underlying hypothesis is very different. Our NeReF encodes the refractive surface normals which deflect the rays spatially, in contrast NeRF models rays' radiance which is in another dimension of spatial domain. 

\subsection{Depth and Normal from Volume Accumulation} 
From NeReF, depth and normal are accumulated along the camera ray. Specifically, given $k$ sample points along a camera ray $\mathbf{r} = \mathbf{o} + \lambda \mathbf{d}$ where $\mathbf{o}$ is the origin of the ray and $\mathbf{d}$ is the ray direction, with each sample point determined by parameter $\lambda_i, i \in \{1,...,k\}$. 
We query the sampling point's volume density $\sigma_i$ and normal $\mathbf{n}_i$. Then, we calculate the surface normal $\mathbf{N}(r_k)$ and depth $D(r_k)$ of the last sampling point $\lambda_k$ as:
\begin{equation}
\begin{split}
    \mathbf{N}(\mathbf{r}_k) = \sum^k_{i=1} \tau_i [1-\exp(-\sigma_i\delta_i) ]\mathbf{n_i} \\
    D(\mathbf{r}_k) = \sum^k_{i=1} \tau_i [1-\exp(-\sigma_i \delta_i)]\sigma_i
\end{split}
\end{equation}
where $\tau_i = \exp\left(-\sum_{j=1}^{i-1} \sigma_j\delta_j\right)$, and $\delta_i = \lambda_{i+1} - \lambda_i$ denotes the distance between two adjacent samples along the ray. Then, the coordinate of 3D point that $\mathbf{r}$ intersects the fluid surface is $\mathbf{p}_s = \mathbf{o} + D(\mathbf{r}_k)\mathbf{d} $

\subsection{Refraction Execution}

The implicit fluid representation should also obey the Snell’s Law when a ray traverse the refractive surface. The refraction process follows Snell's law, i.e., $n_1 \sin \theta_1 = n_2 \sin \theta_2$, where $n_1, n_2$ and $\theta_1, \theta_2$ are the refraction indexes and incident and refracted ray angles respectively.

For each camera ray $\mathbf{r}$, we calculate the refracted ray $\mathbf{r}^\prime = \mathbf{p}_s + \lambda\mathbf{d}^\prime$ using the Snell's law in vector form, as:
\begin{equation}
    \mathbf{d}^\prime = \frac{s \cdot \mathbf{d} + (s a - b)\mathbf{N}(\mathbf{r})}{\|s \cdot \mathbf{d} + (s a - b)\mathbf{N}(\mathbf{r})\|_2}
\end{equation}
where
\begin{equation}
\left\{
\begin{array}{lr}
s = n_1 / n_2, & \\
a = -\mathbf{N}(\mathbf{r}) \cdot \mathbf{d_c}, &\\
b = \sqrt{1-s^2\left(1-(\mathbf{N}(\mathbf{r})\cdot \mathbf{d})^2 \right)}
\end{array}
\right.
\end{equation}

In our case, $n_1$ and $n_2$ are the refractive indexes of air and water, respectively. 

\section{Fluid Surface Reconstruction with NeReF}

We represent the fluid surface implicitly with NeReF. One remaining problem is how to train the network. Previous section describes how to refract a ray using NeReF, in our implementation, we assume the camera ray refracts once when it hits the fluid surface, i.e., $\mathbf{N}(\mathbf{r}) = \mathbf{N}(\mathbf{r}_k)$ for $k\to
 \infty$, but remember this assumption is not required for our NeReF but only for the easy analysis of the fluid surface reconstruction problem. 

We can use the consistency between the refracted rays to optimize the network, but need the ground truth refracted rays. Hence we follow the scheme of previous approaches and place a reference pattern under the water. We first capture images of the pattern without water and then pour the water in. Then we can use the positional differences of corresponding intersection points on pattern between the refracted rays and un-refracted rays and use it a loss for NeReF optimization. The process is visualized in Fig.~\ref{fig:pipeline}.

 \subsection{Problem Formation}

We set the $x-y$ plane of coordinate system to be aligned with the reference pattern, and $z$ direction is up and perpendicular to the pattern plane. Then normal of the reference plane is $\mathbf{n}_{\pi} = [0, 0, 1]$. For a refracted ray $\mathbf{r^\prime}(\lambda) = \mathbf{p}_s + \lambda \mathbf{d}^\prime$, we can obtain the intersection point $\mathbf{q}^\prime$ with the reference plane from ray-plane intersection:

\begin{equation}
    \lambda_\mathbf{q^\prime} = \frac{ - \mathbf{p}_s \cdot \mathbf{n}_{\pi}}{\mathbf{d}^\prime \cdot \mathbf{n}_{\pi}}
\end{equation}
\noindent where $\cdot$ denotes the dot product. Similarly, we can obtain the intersection point $\mathbf{q}$ of the ray without refraction from $\lambda_\mathbf{q}$. Then our aim is to optimize the NeReF network from $\mathbf{q}$ and  $\mathbf{q}^\prime$.

%However, we cannot obtain the ground truth refracted direction to update the network parameters. In order to make the network trainable, we 

%, and a point on it as $\mathbf{p}_{plane}$. If one point $\mathbf{p}$ on the refracted ray $\mathbf{r_{ref}}(\lambda) = \mathbf{p}_s + \lambda\mathbf{d}_r$ intersects the plane, it should satisfy the constraint that 
%\begin{equation}
%     ( \mathbf{r_{ref}}(\lambda) - \mathbf{p}_{plane} ) \cdot %\mathbf{N}_{plane} = 0
%\end{equation}

%Then we can solve the offset $\lambda$ by
%\begin{equation}
%    \lambda = \frac{(\mathbf{p}_{plane}  - \mathbf{p}_s) \cdot \mathbf{N}_{plane}}{\mathbf{d}_r\cdot \mathbf{N}_{plane}}
%\end{equation}

However, there is still one remaining problem that we don't know how to correspond $\mathbf{q}$ and $\mathbf{q}^\prime$. Notice $\mathbf{p}$ and $\mathbf{p}^\prime$ are projection points on the image plane of rays $\mathbf{r}$ and $\mathbf{r}^\prime$. So the ray $\mathbf{r}$ that before refraction can be directly obtained from camera parameters and pixel location $\mathbf{q}$. Then we use the optical flow~\cite{teed2020raft} technique to obtain a dense warp field $\mathcal{W}$ from the distorted frame $I^\prime$ to reference frame $I$, presumably we obtain the shift of $\mathbf{q}^\prime$ as:

\begin{equation}
    I(\mathbf{q}_{gt}) = I^\prime(\mathbf{q}^\prime + d_q)
\end{equation}

\noindent We set $\mathbf{q}_{gt}$ as our ground truth point for the refracted ray $\mathbf{r}^\prime$ of $\mathbf{r}$. Then we refract $\mathbf{r}$ using NeReF network during training and produce a $\mathbf{q_{\mathcal{F}}}$. We define our correspondence loss as the L1 smoothed difference between $\mathbf{q}_{gt}$ and $\mathbf{q_{\mathcal{F}}}$ as:

\begin{equation}
    \mathcal{L}_{corr} = \mathcal{S}^1(\| \mathbf{q_{\mathcal{F}}} - \mathbf{q}_{gt} \|_2)
\end{equation}
\noindent where $\mathcal{S}^1$ denotes the L1 smooth operatoer.

% \subsection{Depth Normal Consistency Loss}
% From NeReF, the depth $D$ and normal $\mathbf{N}$ are recovered from density $\sigma$ and normal $\mathbf{n}_i$ independently. In reality, the depth and normal are correlated as normal can be calculated from derivative of the depth as $\mathbf{N}^\prime (\mathbf{r}) = [\frac{\partial D}{\partial x}, \frac{\partial D}{\partial y}, 1]$. We then normalize $\mathbf{N}^\prime$ and enforce the depth normal correlation using a consistency loss as:

% \begin{equation}
%     \mathcal{L}_{dnc} (D, \mathbf{N}) = \frac{1}{M} \sum_\mathbf{r} \| \mathbf{N}^\prime(\mathbf{r}) - \mathbf{N}(\mathbf{r}) \|_2^2
% \end{equation}
% \noindent where $M$ is the total number of rays. Notice, \cite{thapa2020dynamic} also developed similar loss, but they are trained on a synthetic dataset with ground truth normal. While the ground truth normal is unknown in our case during training.

\subsection{Depth Smoothness}

In practice, we notice depth tends to contain more noise compared to normal. While according to fluid dynamics, depth should be very smooth due to physic constraints. Hence we use a depth smoothness loss:

\begin{equation}
    \mathcal{L}_{ds} (D) = \frac{1}{M} \sum_\mathbf{r} \mathcal{S}^1( | \frac{\partial D(\mathbf{r})}{\partial x}| +  | \frac{\partial D(\mathbf{r})}{\partial y} |)
\end{equation}

Finally, our total loss then is the weighted combination of previous losses as:

\begin{equation}
    \mathcal{L}_{tol} = \mathcal{L}_{corr}  + \lambda_{ds} \mathcal{L}_{ds} 
\end{equation}

Specifically, we set $\lambda_{ds}=0.15$ in our implementation.
\section{Implementation Details}

In our implementation, the NeReF consists of 8 fully connected layers and each layer has 256 channels. The network takes location as input and generates density $\sigma$. Then we use another 3 layers to regress normal from  $\sigma$. Hence the total size of our NeReF is about 4MB.

\noindent \textbf{Training Details.}
We train our models using Adam optimizer with a learning rate 4E-4 which decays 5E-5 per $1000$ iterations during training. Besides, we sample 2048 camera rays for each mini-batch and sample 96 and 192 points from near to far following the hierarchical sampling strategy.
We optimize all our networks on a PC with a single Nvidia GeForce RTX3090 GPU for 10 epochs. The training time is about 2 hours.

\noindent \textbf{System Setup.} To validate the proposition of our NeReF, we construct a fluid capture system consists of a water tank, a pattern underwater and a 10 camera array. The water tank is with size 12 inches for wave simulation. The industrial cameras we use is Z Cam E2 and we place all cameras on on top to record videos of the water. We calibrate the intrinsics and extrinsics of the cameras, and remove lens distortion using OpenCV~\cite{itseez2014theopencv}. We use 9 cameras for NeReF training and the remaining one for evaluation and testing.

%-------------------------------------------------------------------------
\begin{figure*}[t]
\centering
	\centerline{\noindent\includegraphics[width=0.9\linewidth]{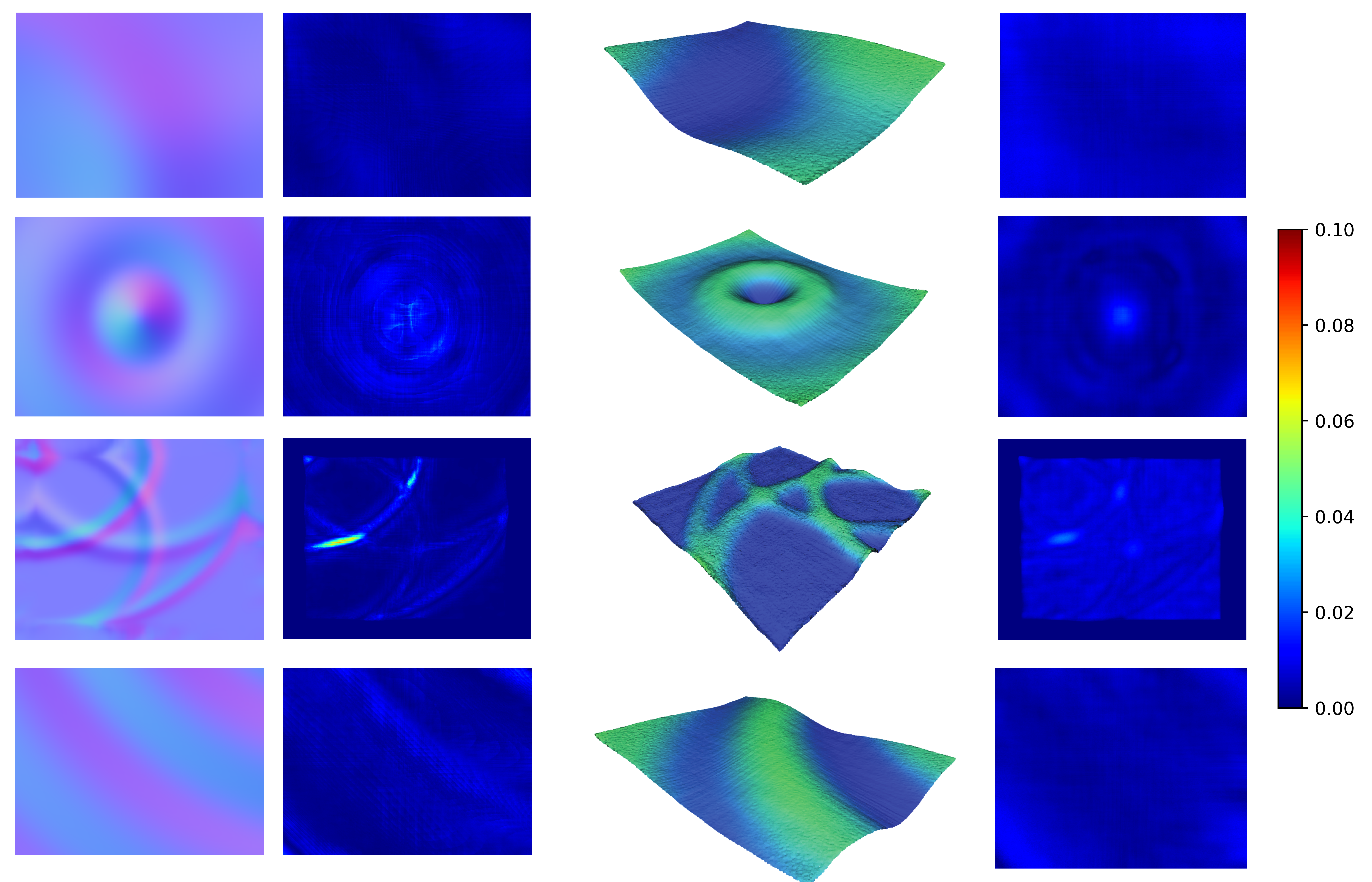} }
	\vspace{-10pt}
\caption{Example fluid surfaces from synthetic data generated by Blender reconstructed by NeReF. From left to right are recovered normal map, normal error, depth visualized as point cloud, and depth error map. The results exhibit very small error and prove the effectiveness of our method.}
\label{fig:gallery}
\end{figure*}

\section{Experiment}

We first evaluate the fluid surface reconstruction approach based on our NeReF, for synthetic fluid data generated by Blender~\cite{Blender}. We exploit the fluid physics simulation ability of Blender and set up a scene that contains 25 cameras and water with size 2 x 1 x 4. We place a binary planar pattern beneath the water, and then use wave modifier to simulate the shallow water equation, 
Grestner’s equation, and Gaussian equation effects. We generate 4 sequences and each sequence contains 90 images. We place 25 pinhole cameras on top of the fluid to observe the pattern distortions under various wave functions. 

% Since our approach relies on the accuracy of the optical flow technique heavy, we fine-tune the RAFT~\cite{teed2020raft} with synthesized water flow distortions. Specifically, we use the pre-trained model of RAFT on the Sintel~\cite{butler2012naturalistic} dataset, and then finetune it with 50000 synthesized fluid images. 
% This way, the optical flow technique adapts to water flow distortion better. 
% Then we use the optical flow~\cite{teed2020raft} technique to obtain a dense warp field $\mathcal{W}$ from the distorted frame $I^\prime$ to reference frame $I
We obtain a dense warp field using the optical flow~\cite{teed2020raft} technique.
Then, we train our NeReF using 9 cameras as describe above and render the depth, normal from NeReF using volumetric integration approach. We compute the relative error, i.e., the error to ground truth ratio, for the depth error and use the L2 norm of the difference between recovered and ground truth normal as the depth and normal error respectively. The recovered normal is smooth and the recovered point cloud is very accurate. We can see errors at most parts of the normal and depth are small, only normal at the peaks of ripple waves show relatively large errors. This is reasonable as the normal changes dramatically in that region. Notice in Fig.~\ref{fig:gallery}, the scale of the error map is very small, i.e., $0.1$, and some normal error in row 3 is just relatively larger.  The synthetic experiments demonstrate that our proposed NeReF works accurately for fluid surface reconstruction.

\subsection{Quantitative Evaluation.}

To quantitatively evaluate our method, we compute the PSNR, LPIPS and SSIM metrics on real data. Recall that we mounted 10 cameras on top of the water tank for capturing the fluid dynamics. However, only 9 of them are used for NeReF training and we use the remaining one for testing. Specifically, after the NeReF is optimized, we synthesize a new view image by setting the sampling camera to be exactly the same with the testing camera location. We compare the synthesized image with the camera image in terms of PSNR, LPIPS and SSIM metrics and the result is shown in Tab.~\ref{tb:quat_compare}. We also include the metrics reported by the original NeRF and for references. We have tried our best to find an alternative fluid surface reconstruction method to compare ours with. We can only Ding'11~\cite{ding2011dynamic} that have the same multi-camera setup with available code. Hence we use Ding'11 approach to recover the depth and normal, and then generate the view and testing camera using ray tracking and report the metrics in Tab.~\ref{tb:quat_compare} too.

\begin{table}[t]
	\centering
	\begin{tabular}{c | c | c | c}
	    \toprule
		\multirow{2}{16pt}{ } & \textbf{PSNR} & \textbf{LPIPS} & \textbf{SSIM} \\ \hline
		 Ding'11 & 25.110 & 0.062 & 0.878 \\ \hline
		 NeRF & 21.560 & 0.217  & 0.684 \\ \hline
		 Ours & 28.926 & 0.033 & 0.942 \\
		\bottomrule
	\end{tabular}
	\rule{0pt}{0.05pt}
	\caption{\textbf{Comparison of our NeReF with the original NeRF and Ding'11~\cite{ding2011dynamic} in term of recover accuracy}. PSNR, LPIPS and SSIM values of our method and the original NeRF on the real fluid sequence.}
\label{tb:quat_compare}
\end{table}

\begin{figure*}
    \centering
    \includegraphics[width=1.0\linewidth]{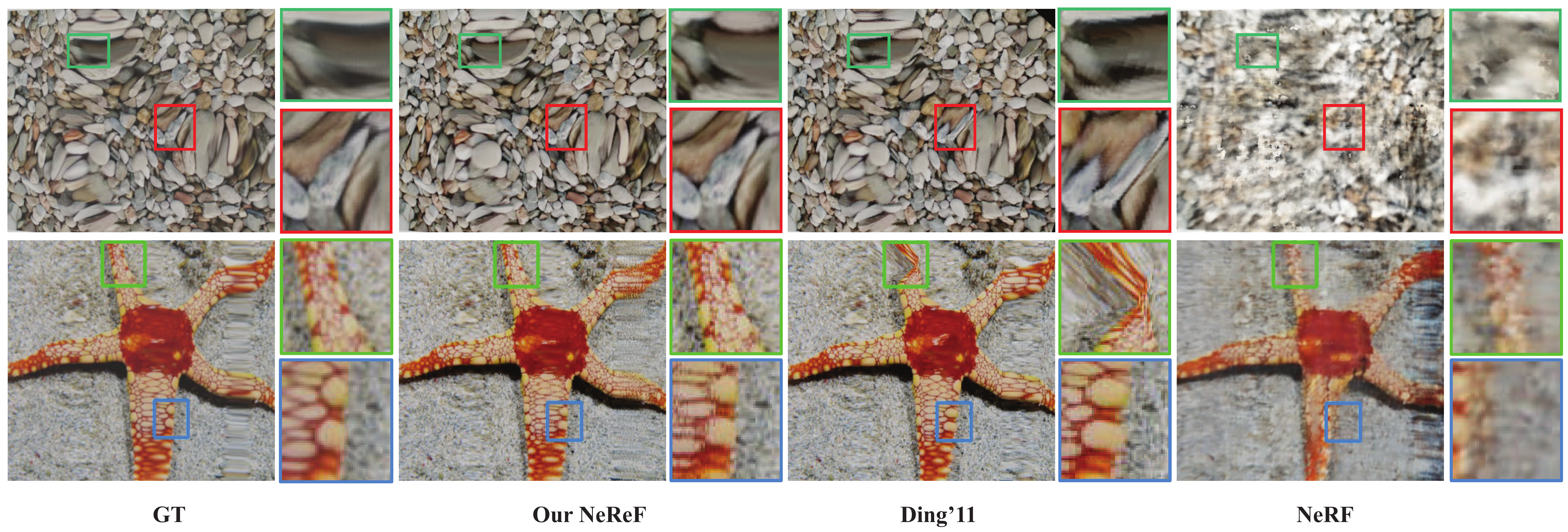}
    \caption{The re-rendering results directly from NeReF (not through ray tracing using depth and normal) on synthetic data compared with other methods.}
    \label{fig:rerender_syn}
\end{figure*}

As we can see the NeReF show good metrics and it means that the synthesized data matches well with the testing camera.

\subsection{Ablation Study}

Here we analysis how several key factors in the system would affect our system:

\textbf{Tolerance of flow error} 

Since optical flow provides the ground truth for our NeReF training, we first evaluate how errors in optical flow affect the final performance. From our synthetic data in above section, we have the ground truth optical flow. Then we add random noise at different amplitude levels, train and test the NeReF model and then use the optimized NeReF model for depth and normal recovery tasks.

\begin{figure}
    \centering
    \includegraphics[width=1.0\linewidth]{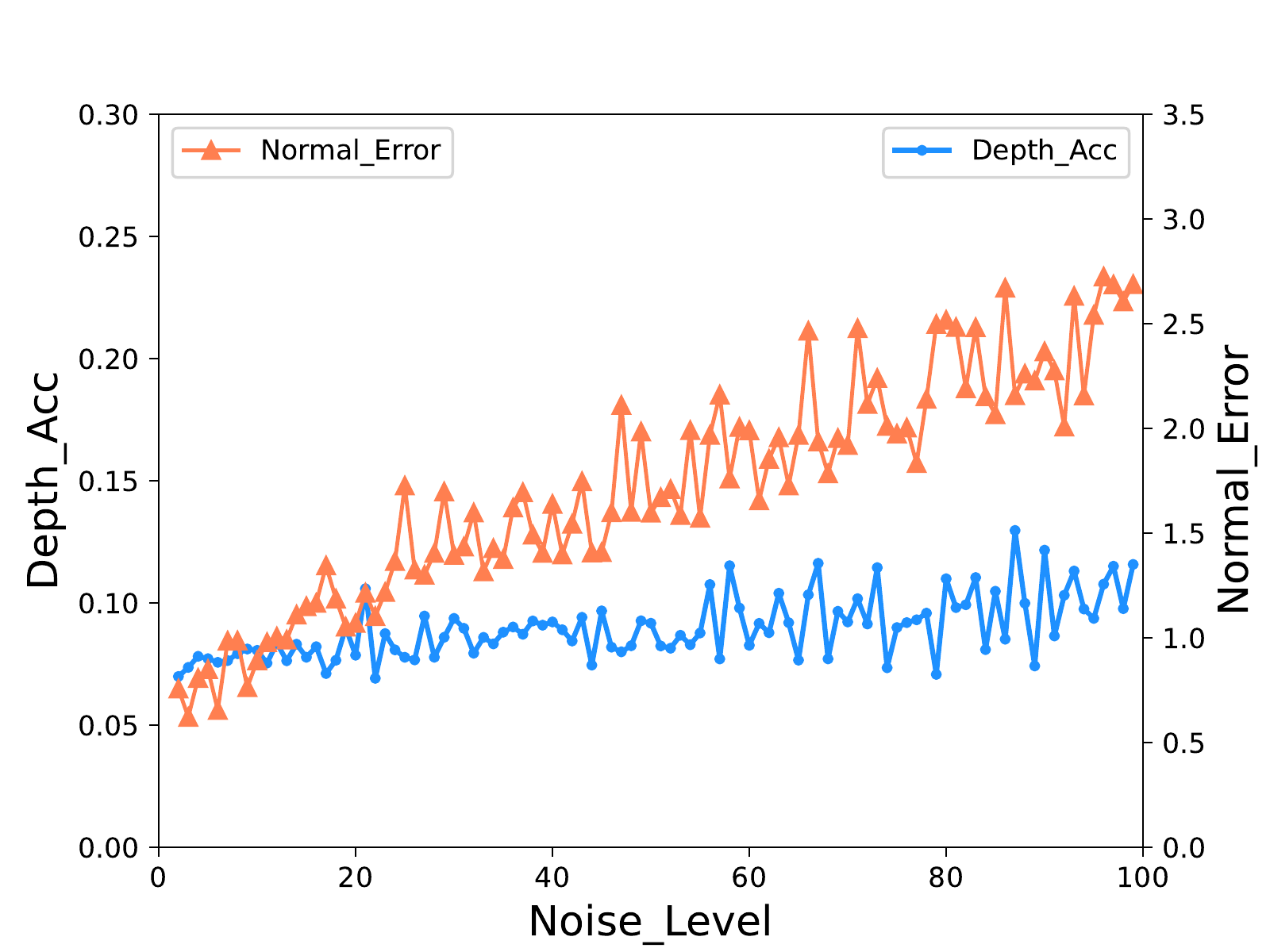}
    \caption{We study how noises in optical flow affect the performance of NeReF. We gradually increase the noise level and train the NeReF for depth and normal recovery, and calculate the error with ground truth.}
    \label{fig:of_noise}
\end{figure}

Figure~\ref{fig:of_noise} shows the depth inaccuracy, which is calculated by error to ground truth depth ratio, and the normal error mean. As we can see, the depth inaccuracy and normal error mean increase when we add larger noise to optical flow. But it also shows that our NeReF is relatively robust to noise in optical flow as the error curve exhibit a linear increasing tendency instead of a exponential increasing curve.

\textbf{Camera Number Evaluation}

The number of cameras is another important term that may affect the performance significantly. We tested the camera numbers from 9, 7, 5 to 3. And we train NeReF with with the corresponding camera views. After training, we synthesize images at testing view points with the trained model and compare with the ground truth depth and normal. We use the average depth and normal error of all testing views. The result is shown in Tab.~\ref{tab:cam_num_study}, as we can see as as the number of cameras decreases, the error in Reconstruction becomes larger, which is as predicted. 

\begin{table}[t]
	\centering
	\begin{tabular}{c | c | c}
	    \toprule
		\multirow{2}{42pt}{Number of cams.} & \textbf{Recovered Depth} & \textbf{Recovered Normal}\\ \cline{2-3}
		   & RMSE (meters) & Angle Diff. (deg.)  \\ \hline
		 9 & 0.02252 & 0.28477  \\ \hline
		 7 & 0.03193 & 0.34128  \\ \hline
		 5 & 0.04540 & 0.43536  \\ \hline
		 3 & 0.05683 & 0.84187  \\ \hline
		\bottomrule
	\end{tabular}
	\rule{0pt}{0.05pt}
	\caption{\label{tab:cam_num_study}\textbf{Ablation study on number of cameras}. RMSE of depth and Average Angle Difference of normal increases as the number of cameras descreases.}
	
\end{table}

%\textbf{Depth Smoothness loss}

\subsection{Refraction and View Synthesis with NeReF}

One important property of NeRF is its ability of synthesizing photo-realitic novel views. This ability comes from the representation of density, and novel view is synthesize via volumetric integration of density and produces depth. Similarly in our NeReF, we can integrate the volume to obtain depth and normal. With the depth and normal, we can directly apply Snell's law to find the refracted ray and hence find where the ray hits after refraction. As such, we can perform background change and novel view synthesis from the NeReF. 

We first compare our NeReF with the original NeRF and ray tacing through depth and normal generated by ~\cite{ding2011dynamic}. The comparison is shown in Fig.~\ref{fig:rerender_syn}. We can see original NeRF performs very poorly as the ray geometry are wrong. Moreover, result from ray tracing~\cite{ding2011dynamic} contains artifacts as ray tracing can only performed at a certain resolution. While the synthesis result from our NeReF is the most close to ground truth.

\begin{figure}
    \centering
    \includegraphics[width=1.0\linewidth]{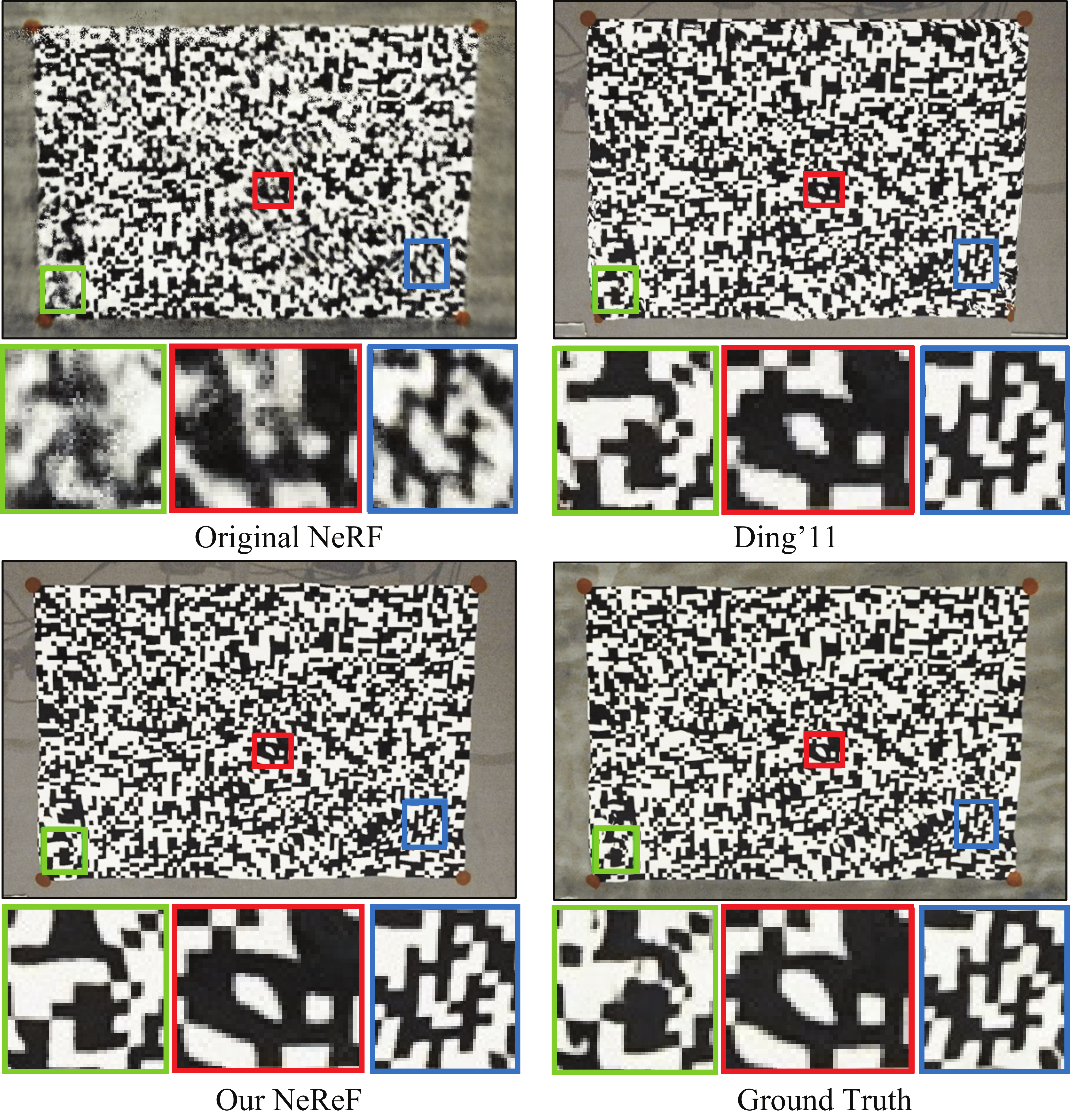}
    \caption{The re-rendering results directly from NeReF (no depth or normal explicitly involved) compared with the original NeRF, ray tracing from Ding'11 depth and normal, the ground truth at the testing view.}
    \label{fig:real_rerenderer}
\end{figure}

Then we compare the novel view synthesis result on real fluid data the fluid data. Fig.~\ref{fig:real_rerenderer} exhibits that results by ray tracing from Ding'11~\cite{ding2011dynamic} depth and normal usually contain minor artifacts, the result from original NeRF is unusable, and results from our approach are very close to the ground truth. The rendered binary pattern is much more clearer and it deforms in the same way as the real image.

\begin{figure}
    \centering
    \includegraphics[width=1.0\linewidth]{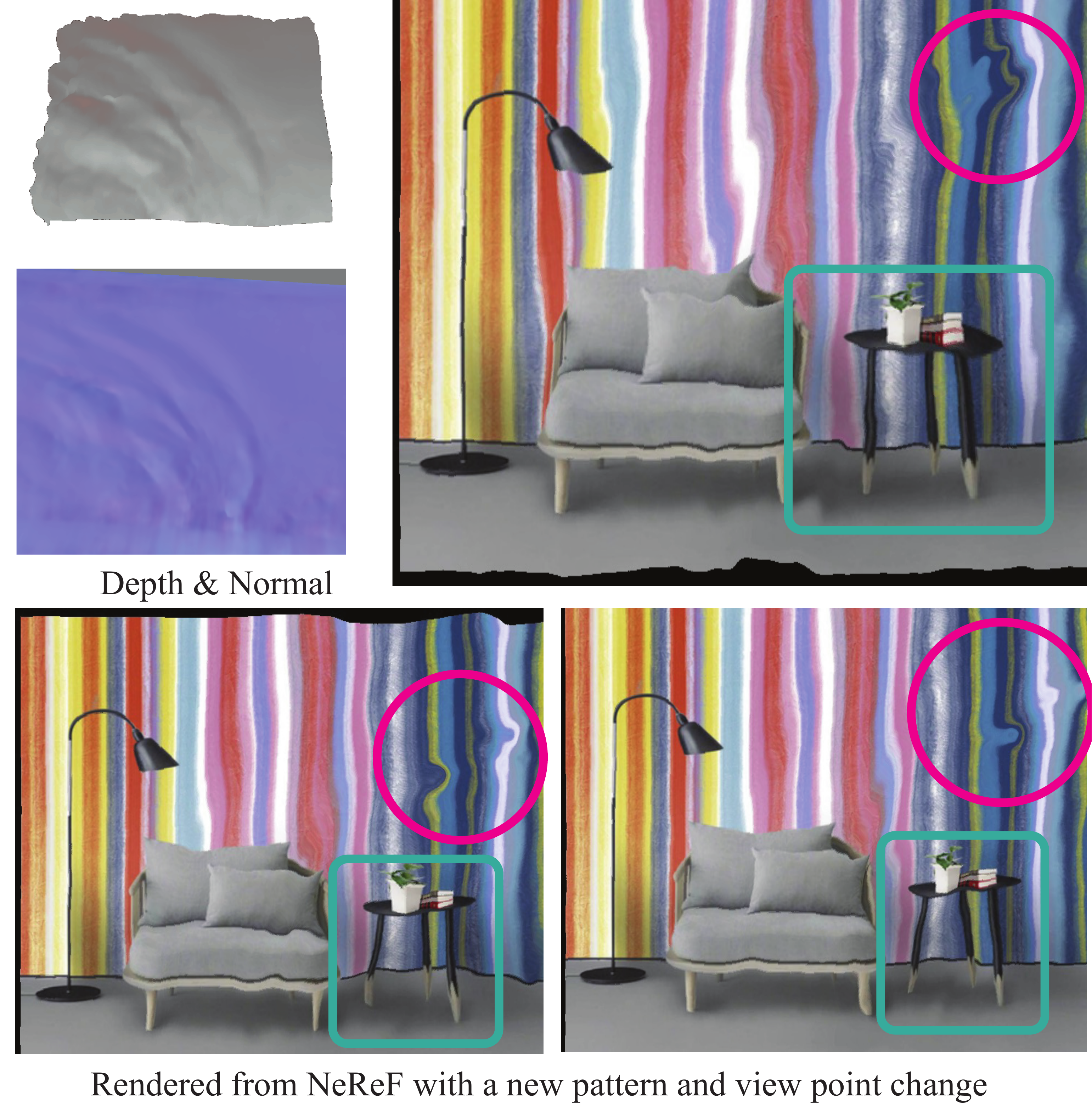}
    \caption{Depth and normal from NeReF optimized on our real captured data. We also show 3 re-rendering image from NeReF with a changed pattern at different viewpoints, notice how distortions are presented at different views.}
    \label{fig:real_depth_normal_render}
\end{figure}

In Fig.~\ref{fig:real_depth_normal_render}, we show the recovered depth and normal for the frame captured in Fig.~\ref{fig:real_rerenderer}. And we re-render use NeReF's volume integration with a new pattern at different viewing points. Notice how distortions are presented differently in each image due to viewing point change.

%-------------------------------------------------------------------------
\section{Discussion}

In this paper, we propose a neural scene representation for refractive fluid surfaces, called the Neural Refractive Field or NeReF. Specifically, we represent the fluid surface as a fully-connected deep network, which takes 3D coordinates as input and output the volume density and normal. We can render the refraction effect directly from the implicit representation.
%We are first to demonstrate that normal can be encoded into a neural field.

We only train and test the NeReF with fluid surface in a water tank. The surface is then rather flat and we assume one time of refraction. This assumption may not hold for water in flow or more complicate fluid surface. Hence, in the future, we plan to employ our method for complicated refractive shape recovery. Moreover, our adaption approach has removed the view dependent radiance from NeRF. Hence the render results from NeReF is a constant color from pattern no matter what direction the target view is. We plan to add the view dependent radiance to NeReF and make it be able to synthesize specular highlights.

%%%%%%%%% REFERENCES
{\small
\bibliographystyle{ieee_fullname}
\bibliography{egbib}
}

\end{document}